\documentclass[journal,10pt,twocolumn,a4]{IEEEtran}
\usepackage{cite}
\usepackage{amsmath,amssymb,amsfonts}
\usepackage{algorithmic}
\usepackage{graphicx}
\usepackage{textcomp}
\usepackage{algorithm}
\usepackage{amsmath,bm}
\usepackage{amssymb}
\usepackage{tabularx}
\usepackage{multirow}
\usepackage{pifont}
\newcommand{\cmark}{\ding{51}}%
\newcommand{\xmark}{\ding{55}}%

\DeclareMathOperator*{\argmin}{argmin}

\newcommand{\tr}{\mbox{tr}\,}

\def\BibTeX{{\rm B\kern-.05em{\sc i\kern-.025em b}\kern-.08em
		T\kern-.1667em\lower.7ex\hbox{E}\kern-.125emX}}

\begin{document}

	\title{Graph Autoencoders for Embedding Learning in Brain Networks and Major Depressive Disorder Identification}
	
	\author{Fuad Noman, Chee-Ming Ting, Hakmook Kang, Rapha\"{e}l C.-W. Phan, Brian D. Boyd, Warren D. Taylor, and Hernando Ombao \vspace{-0.4in}
		\thanks{F. Noman, C.-M. Ting, and R. CW Phan are with the School of Information Technology, Monash University Malaysia, Bandar Sunway, Selangor, 47500 Malaysia (e-mail: fuad.noman@monash.edu; ting.cheeming@monash.edu; raphael.phan@monash.edu). }
		\thanks{H. Kang is with the Department of Biostatistics, Vanderbilt University Medical Center, Nashville, Tennessee 37232 USA (e-mail: h.kang@Vanderbilt.edu).}
		\thanks{W. D. Taylor and B. D. Boyd are with the Center for Cognitive Medicine, Department of Psychiatry, Vanderbilt University Medical Center, Nashville, Tennessee 37212, USA (e-mail: warren.d.taylor@vumc.org; brian.d.boyd@vumc.org).}
		\thanks{H. Ombao, is with Statistics Program, King Abdullah University of Science and Technology, Thuwal, 23955-6900 Saudi Arabia. (e-mail: hernando.ombao@kaust.edu.sa).}}
	
\title{Graph-Regularized Manifold-Aware Conditional Wasserstein GAN for Brain Functional Connectivity Generation}
\author{Yee-Fan Tan, Chee-Ming Ting, Fuad Noman, Rapha\"{e}l C.-W. Phan, and Hernando Ombao \vspace{-0.2in}
\thanks{Y.-F. Tan, F. Noman, R.C.-W. Phan are with the School of Information Technology, Monash University Malaysia, 47500 Malaysia. (e-mails: tan.yeefan@monash.edu; fuad.noman@monash.edu; raphael.phan@monash.edu).} 
\thanks{C.-M. Ting is with the School of Information Technology, Monash
University Malaysia, Subang Jaya 47500, Malaysia, also with the Biostatistics
Group, King Abdullah University of Science and Technology,
Thuwal 23955, Saudi Arabia, and also with the Division of Psychology
and Language Sciences, University College London, London WC1H0AP,
U.K. (e-mail: ting.cheeming@monash.edu).}
\thanks{Hernando Ombao is with the Biostatistics Group, King Abdullah University of Science and Technology, Thuwal 23955, Saudi Arabia (e-mail: hernando.ombao@kaust.edu.sa).}
}

\maketitle

\vspace{-0.05in}
\begin{abstract}
Common measures of brain functional connectivity (FC) including covariance and correlation matrices are semi-positive definite (SPD) matrices residing on a cone-shape Riemannian manifold. Despite its remarkable success for Euclidean-valued data generation, use of standard generative adversarial networks (GANs) to generate manifold-valued FC data neglects its inherent SPD structure and hence the inter-relatedness of edges in real FC. We propose a novel graph-regularized manifold-aware conditional Wasserstein GAN (GR-SPD-GAN) for FC data generation on the SPD manifold that can preserve the global FC structure. Specifically, we optimize a generalized Wasserstein distance between the real and generated SPD data under an adversarial training, conditioned on the class labels. The resulting generator can synthesize new SPD-valued FC matrices associated with different classes of brain networks, e.g., brain disorder or healthy control. Furthermore, we introduce additional population graph-based regularization terms on both the SPD manifold and its tangent space to encourage the generator to respect the inter-subject similarity of FC patterns in the real data. This also helps in avoiding mode collapse and produces more stable GAN training. Evaluated on resting-state functional magnetic resonance imaging (fMRI) data of major depressive disorder (MDD), qualitative and quantitative results show that the proposed GR-SPD-GAN clearly outperforms several state-of-the-art GANs in generating more realistic fMRI-based FC samples. When applied to FC data augmentation for MDD identification, classification models trained on augmented data generated by our approach achieved the largest margin of improvement in classification accuracy among the competing GANs over baselines without data augmentation.
\end{abstract}
\vspace{-0.05in}

\vspace{-0.5cm}
\section{Introduction}
\label{sec:introduction}

\IEEEPARstart{f}{unctional} connectivity (FC) networks composed of interactions among spatially-distinct brain regions are often characterized by cross-correlations of blood-oxygen dependent level signals between different brain regions, measured using functional magnetic resonance imaging (fMRI) \cite{Cassidy2014, Ting2018}. Correlation-based analysis of FC has formed a basis for diverse areas of neuroimaging research, including investigation of neurodevelopmental, psychiatric, and neurological diseases \cite{Greicius2008,Zhang2021}, dynamic connectivity analysis \cite{Hutchison2013}, identification of individuals \cite{Finn2015}, machine learning-based prediction of behavior \cite{Tian2021} and brain disorders \cite{Du2018}. Deep learning (DL) methods have also been introduced recently for identifying brain disorders using fMRI FC patterns, and showed promising improvements over traditional classifiers. These include classification of autism spectrum disorder using autoencoders \cite{Heinsfeld2018}, mild cognitive impairment using convolutional neural networks (CNNs) \cite{Kam2019} and deep Boltzmann machine \cite{Suk2014} to name a few. While most of these works focused on functional connectomic classification, the problem of FC data generation has been less investigated in the state-of-the-art.

FC generation by synthesizing realistic functional connectome profiles associated with healthy and disease groups can have important applications, including data augmentation to improve the performance of FC classifier. In particular, DL models require a large amount of training data to achieve reasonable performance for brain FC classification. This may not be feasible for small-sample settings of fMRI in clinical practice, where classification inevitably encounters problems of over-fitting and difficulty to generalize to unseen samples. Data augmentation has become a powerful technique for classification tasks with limited data, by synthesizing realistic fake data to increase training sample sizes. Generative adversarial networks (GANs) \cite{Ian2014}, a deep generative model for synthetic data generation, offers a novel method for data augmentation in both natural and medical image classifications \cite{Antoniou2017,Frid2018}. GANs have been recently used for neuroimaging data augmentation mainly on raw image synthesis, such as raw MRI \cite{Nguyen2020, Meyer2021} and 3D fMRI brain images \cite{Zhuang2019}. While a few recent work applied GANs to brain structural connectivity augmentation for brain disease classification \cite{Barile2021,Li2021b}, synthetic FC augmentation is largely unexplored. 


Synthesizing FC data is a more challenging problem, as common measures of FC networks including covariance, cross-correlation and precision matrices are symmetric positive definite (SPD) matrices forming a special geometric structure of a cone-shaped Riemannian manifold \cite{Kisung2021, Quang2017}. Since FC matrices lie on a SPD manifold, its elements which represent connecting edges are inherently inter-related. Operations of FC matrices as manifold-valued objects can be better performed based on its corresponding geometric structure of SPD manifold rather than the Euclidean geometry. 
Despite its remarkable success for Euclidean data generation (arrays and grid matrices, e.g., natural images), GANs have been rarely applied for generating manifold-valued data. Distance measures in existing GANs are inappropriate to approximate distance between the true and generated data distributions on manifolds. Direct application of conventional GANs for Euclidean data to generate manifold-valued FC data such as in \cite{Yao2019} disregards the inherent SPD geometry, and thus fails to preserve the global network structure in the generated FC matrices.
In this paper, we target the problem of generating FC network data on the Riemannian manifold of the SPD space, that can take into account of all pairwise FC edges as a whole instead of treating edges independently from each other as in the conventional approaches. Relatively few but increasing number of studies have conducted FC analysis using correlations matrices on the SPD manifold, including computing group-level average and variability \cite{Varoquaux2010}, regression analysis \cite{Deligianni2011}, dynamic FC \cite{Dai2019}, and dimension reduction for machine learning \cite{Qiu2015,Dan2022}. For a review of SPD-based FC analyses see \cite{Kisung2021}. However, no prior work have studied brain FC data generation in the SPD space.

We propose a novel variant of GAN called SPD-GAN for SPD manifold-valued data generation to augment fMRI FC data for brain disorder classification. Our method is inspired by manifold-aware Wasserstein GANs (WGAN) recently introduced in \cite{Huang2019} for manifold-valued image generation. Specifically, it generalizes the Wasserstein distance between the true and generated data distributions of existing WGAN to Riemannian manifold of SPD matrices, and incorporates logarithm and exponential maps in the adversarial loss for data mapping between the manifold and its tangent space. Reconstruction losses are also added to promote similarity between the true and generated data in both the manifold and tangent space. We utilize the affine-invariant Riemannian metric (AIRM), a popular metric for SPD space, to define distance between two SPD objects. Drawing ideas from MotionGAN \cite{Otberdout2020} for motion generation on a hypersphere manifold, we further develop a conditional version of the SPD-GAN that uses class labels to guide generation of SPD-valued data.
Furthermore, we build a graph to encode inter-dependency structure in the SPD data based on geodesic distance on the manifold. Additional graph regularization terms on the SPD manifold and tangent space are then incorporated into the conditional SPD-GAN objective function to enforce the generator to generate SDP data that respect the unique dependency structure of the real data.

The main contributions of this work are as follows:
\begin{itemize}
\item[1)] To our best knowledge, this work is the first to explore a manifold-aware GAN with specialized architecture and new cost functions that operates on SPD manifolds for synthesizing realistic brain FC matrices that can preserve the global dependency structure.
\item[2)] We propose a graph-regularized conditional SPD-GAN (GR-SPD-GAN), a novel extension of SPD-valued WGAN by incorporating conditioned generation and graph regularization. (i) The conditional model enables class-supervised generation of SPD data, which facilitates generation of correlation-based FC matrices according to different experimental groups (diseased and control).
(ii) The graph regularization terms can regularize the SPD-GAN generator to avoid mode collapse and produce more stable GAN training. 
We show empirically that the GR-SPD-GAN can generate better quality synthetic FC data in close resemblance to the true data distribution, and with significant gain in terms of geometry scores compared to its unregularized counterparts. When applied to a population graph where each node represents a subject with associated SPD-valued FC matrix, the proposed graph-regularization approach allows generation of FC matrices that can reflect inter-subject relationships, where similar subjects should share similar connectivity structure.
\item[3)] We demonstrate the usefulness of the generated FC data by using them for data augmentation to enhance brain connectome-based classification. Experimental results on a large resting-state fMRI dataset of major depressive disorder (MDD) show that data augmentation using our GR-SPD-GAN leads to substantial improvement in MDD identification tasks with different downstream FC classifiers, and significantly outperforms several other state-of-the-art GAN-based generators that neglect the SPD geometry in FC data.
\end{itemize}

\section{Background}

\subsection{Generative Adversarial Networks}

The family of GAN techniques \cite{Ian2014} has been very successful for synthesizing natural images. The original GAN training is formulated as a min-max game between two competing networks: the generator $G$ and the discriminator $D$. The $G$ maps the random noise into synthetic data approximating the real data, while the $D$ learns to discriminate between the real and the generated data. Theoretically, this framework minimizes the Jensen-Shannon divergence between the distributions of true data and generated samples. The state-of-the-art GANs like Wasserstein GAN (WGAN) \cite{Martin2017} minimizes the Wasserstein-1 distance between the real and synthetic data distributions. An improved WGAN with gradient penalty (WGAN-GP) \cite{Ishaan2017} was proposed to produce more stable GAN training by penalizing the norm of the discriminator with respect to its input. Consider a dataset of examples $\bm{x}_1, \ldots, \bm{x}_M$ sampled from a real data distribution $\mathbb{P}_r$. In WGAN-GP, $G$ and $D$ are trained by solving the following minimax problem
 \begin{equation*}
\label{eqn:wgangp}
    \begin{aligned}
        \min_G \max_D L(G,D)= & \mathbb{E}_{{\bm x} \sim \mathbb{P}_r}[D(\bm x)] - \mathbb{E}_{G(\bm z) \sim \mathbb{P}_g}[D(G(\bm z))]  \\
        &+ \lambda \: \mathbb{E}_{\bm \hat{{\bm x}} \sim \mathbb{P}_{\bm \hat{{\bm x}}}}[(\Vert \nabla_{\hat{\bm x}} D(\hat{\bm x}) \Vert_2 - 1 )^2 ], 
    \end{aligned}
\end{equation*}
where $\bm z$ is random noise, $\mathbb{P}_g$ is the distribution of generated samples, and $\hat{\bm x}$ is random sample following distribution $\mathbb{P}_{\bm \hat{{\bm x}}}$ that is sampled along straight lines between pairs of points from $\mathbb{P}_r$ and $\mathbb{P}_g$, $\nabla_{\hat{\bm x}} D(\hat{\bm x})$ is the gradient with respect to $\hat{\bm x}$, and $\lambda$ is the weighted coefficient of the gradient penalty.

\vspace{-0.05in}
\subsection{The Geometry of SPD Manifolds}


Let ${\bm x}$ be an $n \times n$ real SPD matrix which satisfies the property that ${\bm \alpha}^T {\bm x} {\bm \alpha} > 0$ for all non-zero ${\bm \alpha} \in \mathbb{R}^n$. The space of $n \times n$ SPD matrices, denoted by $\mathcal{S}^{++}(n)$ is a Riemannian manifold $\mathcal{M} = \mathcal{S}^{++}(n)$. A Riemannian manifold $\mathcal{M}$ is a smooth manifold equipped with a Riemannian metric for defining distances on the manifold. Geometrically, a tangent vector is a vector that is tangent to the manifold at a given point ${\bm y} \in \mathcal{M}$. Let $T_{\bm{y}}\mathcal{M}$ denote the tangent space of $\mathcal{M}$ at ${\bm y}$, i.e., the set of all tangent vectors at ${\bm y} \in \mathcal{M}$ which constitutes an Euclidean space. A Riemannian metric on $\mathcal{M}$ is a family of inner product ${\langle \cdot, \cdot \rangle}_{\bm y}: T_{\bm y}\mathcal{M} \times T_{\bm y}\mathcal{M} \rightarrow \mathbb{R}$ defined on each $T_{\bm y}\mathcal{M}$ that varies smoothly with the base point ${\bm y}$. SPD matrices $\mathcal{S}^{++}(n)$ are most widely studied when endowed with the affine-invariant Riemannian metric (AIRM) defined as
\begin{align}
\langle {\bm v}, {\bm w}\rangle_{\bm{y}} & = \langle \bm{y}^{-1/2} {\bm v} \bm{y}^{-1/2}, \bm{y}^{-1/2} {\bm w} \bm{y}^{-1/2}\rangle_F \notag \\ 
& = \tr(\bm{y}^{-1/2}{\bm v}\bm{y}^{-1/2}{\bm w}),
\end{align}
where ${\bm v}, {\bm w} \in T_{\bm y}\mathcal{M}$ are two tangent vectors at point ${\bm y} \in \mathcal{M}$. For SPD manifold $\mathcal{M} = \mathcal{S}^{++}(n)$, $T_{\bm y}\mathcal{M} \cong \mathcal{S}(n)$ where $\mathcal{S}(n)$ is the vector space of $n \times n$ symmetric matrices. The AIRM has several useful properties such as invariance to affine transformation and matrix inversion.

\begin{figure}[!t]
\centerline{\includegraphics[width=0.8\columnwidth]{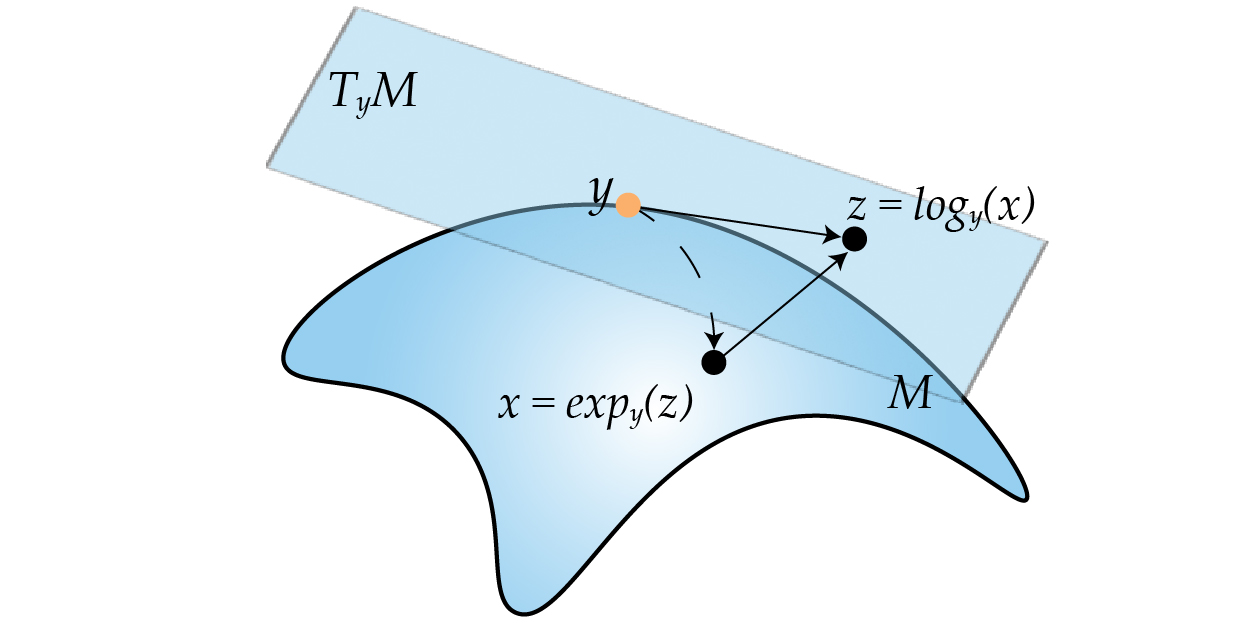}}
\vspace{-0.2cm}
\caption{Exponential and logarithm maps on a Riemannian manifold \cite{Quang2017}.}
\label{riemannian_geometry}
\vspace{-0.1in}
\end{figure}

There are two main operations that connect the manifold $\mathcal{M}$ and the tangent plane $T_{\bm y}\mathcal{M}$: (1) Riemannian exponential map at point ${\bm y}$, $\exp_{\bm y}: T_{\bm y}\mathcal{M} \rightarrow \mathcal{M}$, which projects a tangent vector ${\bm v}$ to a point in $\mathcal{M}$
\begin{equation} \label{eqn:airm_exp}
\exp_{\bm y}({\bm v}) = {\bm y}^{-1/2}\text{Exp}( {\bm y}^{-1/2} {\bm v} {\bm y}^{-1/2})  {\bm y}^{-1/2}, \ \ {\bm v}\in\mathcal{S}(n), 
\end{equation}
where $\text{Exp}$ denotes the matrix exponential. (2) Riemannian logarithmic map at ${\bm y}$, $\log_{\bm y}: \mathcal{M} \rightarrow T_{\bm y}\mathcal{M}$, which maps any SPD matrix ${\bm x} \in \mathcal{M}$ to its tangent space
\begin{equation}
\label{eqn:airm_log}
\log_{\bm y}({\bm x}) = {\bm y}^{-1/2}\text{Log}( {\bm y}^{-1/2} {\bm x} {\bm y}^{-1/2})  {\bm y}^{-1/2}, \ \ {\bm x}\in\mathcal{S}^{++}(n),
\end{equation}
where $\text{Log}$ denotes the principal matrix logarithm. The geometrical interpretation of the exponential and logarithm maps on $\mathcal{M}$ is shown in Fig.~\ref{riemannian_geometry}. Under AIRM, the geodesic distance between two SPD matrices ${\bm x},{\bm y} \in \mathcal{M} = \mathcal{S}^{++}(n)$ follows
\begin{equation}
\label{eqn:airm_geodesic}
d({\bm x}, {\bm y}) =\| \text{Log} ({\bm x}^{-1/2} {\bm y} {\bm x}^{-1/2} )\|_F,
\end{equation}
where $\|\cdot\|_F$ denotes the Frobenius norm.

\begin{figure*}[!htb]
\vspace{-0.2cm}
\centerline{\includegraphics[width=\textwidth]{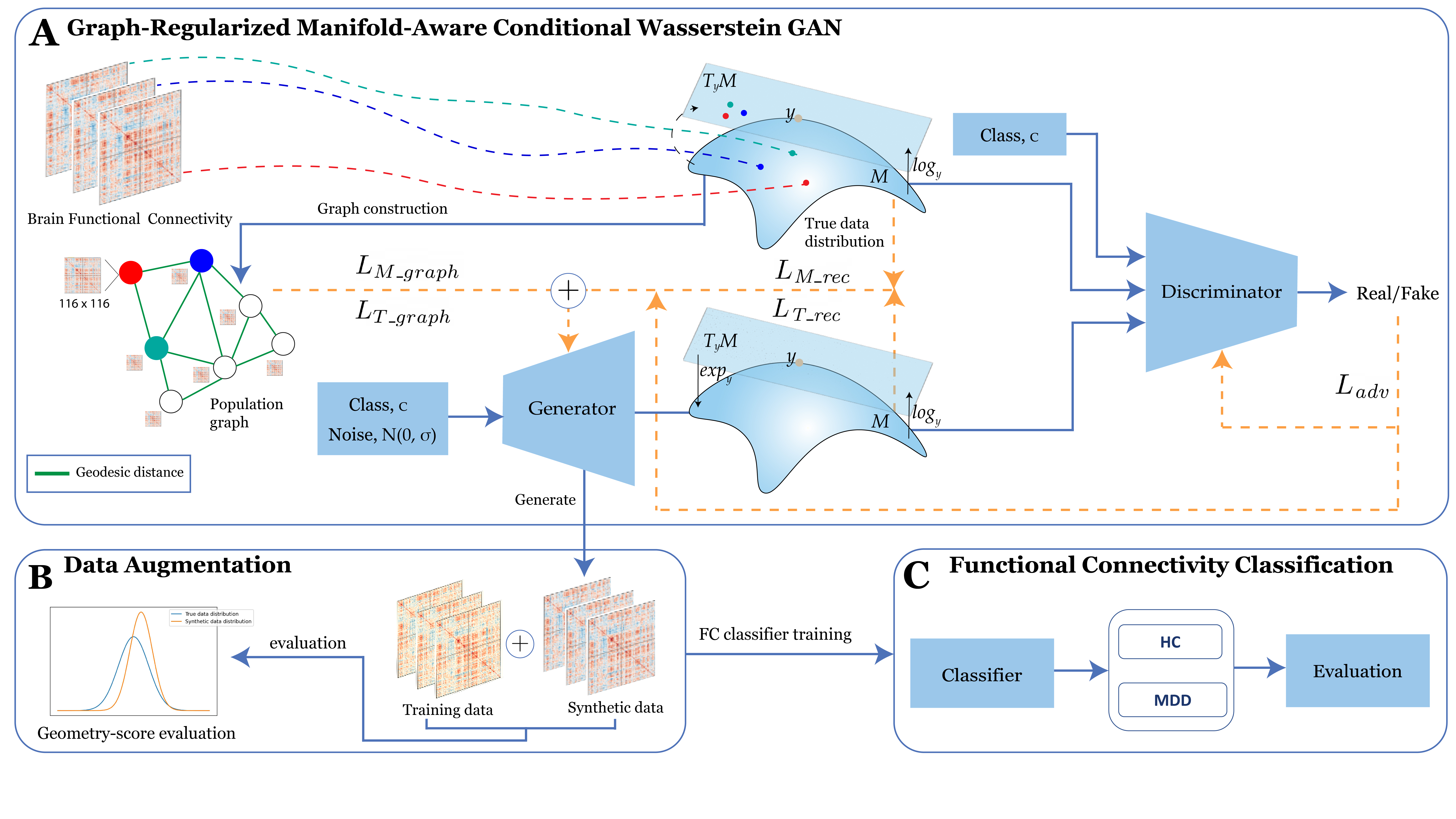}}
\vspace{-0.8cm}
\caption{Overview of the proposed framework: (a) GR-SPD-GAN is trained on a Riemannian manifold using $log$ and $exp$ operations. Population graph regularizer is added to encourage inter-subject similarities during the training. (b) Trained generator of GR-SPD-GAN is used to generate synthetic data. Geometry score is used to evaluate the synthetic dataset. (c) Synthetic data is used in data augmentation for FC classification.}
\label{fig:overview}
\vspace{-0.1in}
\end{figure*}

\section{Methods}

Fig.~\ref{fig:overview} shows an overview of the proposed graph-regularized manifold-aware conditional Wasserstein GAN (GR-SPD-GAN) for SPD-matrix-valued data generation with an application to synthetic FC data augmentation for brain disorder classification. The proposed framework consists of three stages. (a) Model Training: FC matrices estimated from fMRI are represented as compact points on the SPD manifold. The GR-SPD-GAN is trained to learn the distribution of the SPD-valued FC data associated to each class (brain disorder or healthy control group). Riemannian logarithm and exponential maps are exploited to minimize the Wasserstein distance between the distributions of real and generated data under an adversarial training with a generator and a discriminator. A population graph that encodes inter-subject similarity in FC structure is constructed based on geodesic distance on the SPD space. It is then used as a regularizer at the generator to force the generated FC data to respect the inter-subject dependency in the real data. (b) Data augmentation: After training, we use the resulting generator to generate synthetic FC data as new points on the SPD manifold conditioned on the experimental class. The generative performance is measured by the geometric score that compares the geometrical properties of the underlying data manifold and the generated one. (c) FC-based classification: The generated FC data is used to augment training data for subsequent brain disorder classification.

\vspace{-0.05in}
\subsection{Manifold-Aware WGAN for SPD Matrices}

\subsubsection{SPD-GAN Network}

Consider a set of $M$ training samples $\bm X = \{\bm x_i, c_i\}^M_{i=1}$, where $\bm x_i$ represents the FC matrix of subject $i$, and $c_i$ indicates the corresponding class label of the experimental group (i.e., $c_i=0$: healthy control; $c_i=1$: a certain brain disorder). The FC matrix can be estimated by cross-correlations between fMRI time series extracted over voxels and brain regions of interest (ROIs). 
Our aim is to design a mapping function that converts random noise to synthetic FCs given the class $c$. FC matrices $\bm x_i \in \mathcal{M}$ (e.g., covariance and correlation matrices) are SPD data residing on an SPD manifold $\mathcal{M} = \mathcal{S}^{++}(n)$. Inspired by the manifold-aware GANs \cite{Huang2019,Otberdout2020} for image and motion generation, we propose a SPD manifold-aware WGAN (SPD-GAN) to exploit the geometry of the SPD manifold to learn the generative distribution of FCs associated with each experimental group. Analogouus to MotionGAN \cite{Otberdout2020} for motion generation on a hypersphere manifold, our SPD-GAN is a conditional version of manifold-aware WGAN by \cite{Huang2019}, but generates new points in the SDP manifold. 

The proposed SDP-GAN architecture is shown in Fig.~\ref{fig:overview}A. It consists of two adversarial networks: (1) A generative model $G:\mathbb{R}^n \rightarrow \mathcal{S}^{++}(n)$ that maps an $n$-dimensional noise vector $\bm z$ sampled from a prior distribution $\mathbb{P}_{\bm z}$ to a synthetic SDP matrix $\tilde{\bm{x}} \in \mathcal{S}^{++}(n)$ from a manifold-valued generating distribution $\mathbb{P}_g$. (2) A discriminative model $D$ that estimates the probability of a given input $\bm{x} \in \mathcal{S}^{++}(n)$ being sampled from the real data distribution $\mathbb{P}_r$ (from the training data) rather than the generating distribution $\mathbb{P}_g$ (from the $G$).
Both $G$ and $D$ are conditioned on the class label $c$. This conditioning is performed by incorporating $c$ as the input to the $G$ and $D$. To generate valid SPD data from the generative network $G$, we employ the Riemannian exponential map (\ref{eqn:airm_exp}) at a reference point $\bm{y}$ to transfer the output of $G$, which is a symmetric matrix $G({\bm z},c) \in \mathcal{S}(n)$, to the SPD manifold. The logarithm map (\ref{eqn:airm_log}) is used to project the SPD matrices (the real data $\bm{x}$ and the generated data $\tilde{\bm{x}} = \exp_{\bm y}(G({\bm z},c))$) to the tangent space $T_{\bm y}(\mathcal{S}^{++}(n)) \cong \mathcal{S}(n)$ at $\bm y$, before being presented as inputs to the discriminator network $D$.
Since the tangent space is a vector space, any regular neural networks designed for Euclidean data can be used as $D$.

\subsubsection{Loss Function}

To train the $G$ and $D$ of the SPD-GAN, we propose the following objective function, which is a weighted sum of the adversarial loss $\mathcal{L}_{adv}$, the reconstruction loss $\mathcal{L}_{M\_rec}$ in the SPD manifold $\mathcal{S}^{++}(n)$, and the reconstruction loss $\mathcal{L}_{T\_rec}$ in the tangent space $T_{\bm y}(\mathcal{S}^{++}(n))$
\begin{multline}
\label{eqn:loss-SPDGAN}
\min_G \max_D \mathcal{L}_{\textit{SPD-GAN}}(G,D) = \alpha_1 \mathcal{L}_{\textit{adv}}(G,D) + \\ \alpha_2 \mathcal{L}_{\textit{M\_rec}}(G) + \alpha_3 \mathcal{L}_{\textit{T\_rec}}(G).
\end{multline}
The adversarial loss $\mathcal{L}_{adv}$ generalizes the objective function of WGAN-GP \cite{Ishaan2017} to the SPD-manifold valued data, and is a conditional version of the objective function in \cite{Huang2019}.

\begin{equation}
\begin{aligned}
\label{eqn:adv_loss}
\mathcal{L}_{\textit{adv}}(G,D) =  &~ \mathbb{E}_{{\bm x} \sim \mathbb{P}_r}\left[D\left(\log _{{\bm y}}({\bm x}), c \right)\right] \\
&-\mathbb{E}_{G(\bm z) \sim \mathbb{P}_g}\left[D\left(\log _{{\bm y}}\left(\exp _{{\bm y}}(G({\bm z}, c))\right)\right]\right.\\
&+\lambda \mathbb{E}_{\bm \hat{{\bm x}} \sim \mathbb{P}_{\bm \hat{{\bm x}}}}\left[\left(\left\|\nabla_{\bm \hat{{\bm x}}} D(\bm \hat{{\bm x}}, c)\right\|_{2}-1\right)^{2}\right],
\end{aligned}
\end{equation}
where $\exp_{\bm y}(\cdot)$ and $\log_{\bm y}(\cdot)$ are the Riemannian logarithm and exponential maps defined for the SPD manifold at a particular point ${\bm y}$, in \eqref{eqn:airm_exp} and \eqref{eqn:airm_log}, respectively. The first two terms denote the estimate of the Wasserstein distance $W(\mathbb{P}_r,\mathbb{P}_g)$ between the real and generated SPD data that the generator $G$ learns to minimize. The last term represents the gradient penalty as in \cite{Ishaan2017} defined for the manifold-valued data, where $\hat{\bm x}$ is sampled uniformly along straight lines between pairs of points sampled from the real data distribution $\mathbb{P}_r$ and the generating distribution $\mathbb{P}_g$
\begin{equation}
\label{eqn:interpolated_dist}
\hat{\bm x} = (1 - \epsilon) \log_{\bm y}(\bm x) + \epsilon \log_{\bm y}(\exp_{\bm y}(G(\bm z, c)))),
\end{equation}
with $0 \leq \epsilon \leq 1$, and $\nabla_{\bm \hat{{\bm x}}}D(\bm \hat{{\bm x}}, c)$ is the gradient with respect to $\hat{\bm x}$.

We incorporate two reconstruction losses $\mathcal{L}_{M\_rec}$ and $\mathcal{L}_{T\_rec}$ to further encourage the generator to synthesize data approximating the real one on both the SPD manifold and its tangent space, respectively. $\mathcal{L}_{M\_rec}$ measures the distances between the generated samples $\tilde{\bm{x}} = \exp_{\bm y}(G(\bm z,c))$ and their corresponding ground-truth real data $\bm x$ on the manifold $\mathcal{S}^{++}(n)$
\begin{equation}
\label{eqn:m_rec}
\mathcal{L}_{\textit{M\_rec}}(G) =  \mathbb{E}_{{\bm x} \sim \mathbb{P}_r} \mathbb{E}_{G(\bm z) \sim \mathbb{P}_g} d^2(\exp_{\bm y}(G(\bm z, c)), \bm x),\\
\end{equation}
where $d^2(\cdot,\cdot)$ is the squared geodesic distance under AIRM defined in \eqref{eqn:airm_geodesic}. $\mathcal{L}_{T\_rec}$ quantifies the similarities between the generated tangent vectors $\log_{\bm y}(\exp_{\bm y}(G(\bm z,c))) \in T_{\bm y}(\mathcal{S}^{++}(n))$ and their associated ground-truth vectors $\log_{\bm y}(\bm x)$ projected on $T_{\bm y}(\mathcal{S}^{++}(n))$
\begin{equation}
\label{eqn:t_rec}
\mathcal{L}_{T\_rec}(G) = \mathbb{E}_{{\bm x} \sim \mathbb{P}_r} \mathbb{E}_{G(\bm z) \sim \mathbb{P}_g} \Vert \log_{\bm y}(\exp_{\bm y}(G(\bm z, c)) - \log_{\bm y} (\bm x)) \Vert_1,
\end{equation}
where $\Vert \cdot \Vert_1$ denotes the $\mathcal{L}_1$-norm. The generator is trained to minimize the two reconstruction losses \eqref{eqn:m_rec} and \eqref{eqn:t_rec}. To define the tangent space $T_{\bm y}(\mathcal{S}^{++}(n))$ used in the training of our SPD-GAN, we exploit the Fréchet mean \cite{Simone2009, Hermann1977} of the training data $\bm X = \{\bm x_i\}^M_{i=1}$ to define the reference point ${\bm y}$ as
\begin{equation}
\begin{aligned}
\label{eqn:frechet_mean}
{\bm y} = \argmin_{{\bm x} \in \mathcal{M}} \sum^{M}_{i=1} d^2(\bm x, \bm x_i)
\end{aligned}
\end{equation}
which is the minimizer of average squared geodesic distances
over the training FC data.

\vspace{-0.05in}
\subsection{Graph-Regularized SPD-GAN}

Brain FC matrices tend to exhibit similar or highly-correlated patterns between subjects sharing similar demographics, phenotypes or clinical conditions. To incorporate the relationships between subjects, we introduce a novel graph-based regularizer in SPD manifold at the generator of the SPD-GAN to enforce the generated data $\tilde{\bm{x}}$ to respect the inter-subject FC similarity in the real data $\bm{x}$.

\subsubsection{Graph Construction}
Graphs provide a natural way of representing a population of subjects and their relationships. Let $\mathcal{G} = (\mathcal{V},\mathcal{E},{\bf W})$ be an undirected weighted population graph, where $\mathcal{V}$ is the node set (each node represents a subject), $\mathcal{E}$ is the edge set specified by $(i,j, w_{ij})$, $i,j \in \mathcal{V}$ representing connections between subjects, and ${\bf W} = [w_{ij}]$ is the $M \times M$ weighted adjacency matrix. The weight $w_{ij}$ of an edge provides a measure of similarity between subjects $i$ and $j$. The set of SPD-matrix valued FC data $\bm{x}_1, \ldots, \bm{x}_M \in \mathcal{S}^{++}(n)$ can be considered as signals defined on the graph $\mathcal{G}$.

The graph-based inter-subject relationships can be computed using the Euclidean distance between vectorized FC matrices of pairs of subjects, given by
\begin{equation} \label{eqn:pop_graph_euclidean}
w_{ij} = \exp\left(\frac{-\Vert \text{vec}(\bm x_i) - \text{vec}(\bm x_j) \Vert^2}{\sigma^2}\right),
\end{equation}
where the Gaussian heat kernel \cite{Li2021a} is adopted, and $\sigma$ is the kernel scalar parameter. However, \eqref{eqn:pop_graph_euclidean} neglects the geometric properties of SPD manifold. We exploit the geodesic distance between FCs to define the similarity between subjects
\begin{equation}
\label{eqn:pop_graph_manifold}
w_{ij} = \exp\left(\frac{-d^2(\bm x_i, \bm x_j)}{\sigma^2}\right).
\end{equation}
Note that the weights decay with distance. Specifically, $w_{ij}$ assumes a high value if $\bm x_i$ and $\bm x_j$ are similar, and a small value if they are different.

\subsubsection{Graph-based regularization in SPD manifold}

To encourage the SPD-GAN generator to generate FC matrices that respect the inter-subject similarity in the real data, we further introduce two additional graph regularization terms, on both the SPD manifold $\mathcal{S}^{++}(n)$ and its tangent space $T_{\bm y}(\mathcal{S}^{++}(n))$ at ${\bm y}$, defined respectively by
\begin{align}
\mathcal{L}_{M\_graph}(G) &= \sum_i \sum_j { w_{ij} d^2(\tilde{\bm{x}}_i, \tilde{\bm{x}}_j)}, \label{eqn:m_graph} \\
\mathcal{L}_{T\_graph}(G) &= \sum_i \sum_j { w_{ij} \Vert \tilde{\bm{v}}_i - \tilde{\bm{v}}_j \Vert}_F^2, \label{eqn:t_graph}
\end{align}
where $\tilde{\bm{x}}$ and $\tilde{\bm{v}} = \log_{\bm y}(\exp_{\bm y}(G(\bm z,c)))$ are the generated SPD data and associated tangent vectors, respectively. Note that in $\mathcal{L}_{M\_graph}$, geodesic distance is used to define the pairwise differences in the generated SPD matrices $\tilde{\bm{x}}_i$ and $\tilde{\bm{x}}_j$ over connected subjects $i$ and $j$. By minimizing the objective functions \eqref{eqn:m_graph} and \eqref{eqn:t_graph}, it enforces the generator to synthesize similar (or smooth) SPD matrices for subjects with high similarity in brain FC structure, as defined over the population graph $\mathcal{G}$ of the real data.

Incorporating the graph regularizers \eqref{eqn:m_graph} and \eqref{eqn:t_graph} into the objective function \eqref{eqn:loss-SPDGAN} of SPD-GAN gives a new objective of the graph-regularized SPD-GAN (GR-SPD-GAN) as follows
\begin{multline}
\label{eqn:loss-GR-SPDGAN}
\min_G \max_D \mathcal{L}_{\textit{GR-SPD-GAN}}(G,D) = \alpha_1 \mathcal{L}_{\textit{adv}}(G,D) + \alpha_2 \mathcal{L}_{\textit{M\_rec}}(G) \\ + \alpha_3 \mathcal{L}_{\textit{T\_rec}}(G) + \alpha_4 \mathcal{L}_{M\_graph}(G) + \alpha_5 \mathcal{L}_{T\_graph}(G).
\end{multline}

During the training, the generator $G$ generates a matrix $\tilde{\bm x}$, where $\exp$ and $\log$ operations are applied to enforce the generated output to be projected on the desired manifold $\mathcal{M}$ and tangent space $T_{\bm{y}}\mathcal{M}$ based on $\bm y$. The discriminator $D$ takes the tangent vector $ \bm v = \log_{\bm y}(\bm x)$, and generated tangent vector $ \bm \tilde{v} = \exp_{\bm y}(\log_{\bm y}(G(\bm z, c)))$ as input; to score the probability of a data belonging to the true data distribution $\mathbb{P}_r$. Population graph is applied to impose regularization on generator $G$ to generate close FCs resemblance on both manifold $\mathcal{M}$ and tangent space $T_{\bm{y}}\mathcal{M}$. As such, the generator $G$ learns the real data distribution and synthesizes similar data. The proposed graph regularization of the generator can improve the GAN training, by avoiding mode collapse and producing stable training. This can result in significantly better quality of the generated SPD data compared to the unregularized SDP-GAN, as measured in terms of geometric score. Algorithm ~\ref{alg:grmacwg} shows the training steps of GR-SPD-GAN. Both the generator $G$ and discriminator $D$ are implemented via neural networks, parametrized with network parameters $\bm{\theta}$ and $\bm{u}$, respectively. The details of the network architecture for $G$ and $D$ are described in Section IV.B. During the training process, the network parameters of $D$ and $G$ are updated alternately by fixing one of them.

\begin{algorithm}[!t]
\caption{Training of the proposed GR-SPD-GAN}\label{alg:grmacwg}
\begin{algorithmic}[1]
\REQUIRE  $\bm{x}_i$, training data with their corresponding labels ${c_i}$; \bm{$\theta_0$}, initial generator parameters; \bm{$u_0$}, initial discriminator parameters; $M_b$, batch size; $\alpha$, learning rate, $w_{ij}$, weights of constructed population-graph; $n_{d}$, the discriminator iterations per generation iteration; $n_g$, generator update iterations.
\FOR{n=0, ..., $n_{g}$}
\FOR{t=0, ..., $n_{d}$}
\STATE Sample $\{\bm{x}_i, c_i\}^{M_b}_{i=1} \sim \mathbb{P}_r$ a batch of examples from real data distribution
\STATE Sample $\{\bm z_i\}^{M_b}_{i=1} \sim \mathbb{P}_z$ a batch of noise samples from noise prior
\STATE Compute stochastic gradient $\nabla_{\bm u}$ of Eq\eqref{eqn:loss-GR-SPDGAN} with respect to $\bm u$
\STATE Update $\bm u \leftarrow \bm u + \alpha \cdot $AdamOptimizer($\bm u, \nabla_{\bm u}$)
\ENDFOR
\STATE Sample $\{\bm{x}_i, c_i\}^{M_b}_{i=1} \sim \mathbb{P}_r$ a batch of examples from real data distribution
\STATE Sample $\{\bm z_i\}^{M_b}_{i=1} \sim \mathbb{P}_z$ a batch of noise samples from noise prior
\STATE Compute stochastic gradient $\nabla_{\bm \theta}$ 
of Eq\eqref{eqn:loss-GR-SPDGAN} with respect to $\bm \theta$
\par
\STATE Update $\bm \theta \leftarrow \bm \theta + \alpha \cdot $AdamOptimizer($\bm \theta, \nabla_{\bm \theta}$)
\ENDFOR
\end{algorithmic}
\end{algorithm}


\vspace{-0.05in}
\subsection{Data Augmentation for FC Classification}

After training, we apply the learned generator $G$ of GR-SPD-GAN to generate synthetic FC matrices for data augmentation in brain disorder classification. The generation steps are outlined in Algorithm \ref{alg:fca}. The generator $G$ receives input from the noise prior $\mathbb{P}_z \sim \mathcal{N}(0,1)$ to generate a point on the tangent space conditioned on each class of the brain networks, which is then transformed by using the exponential map to its corresponding point on the SPD manifold, which represents a FC matrix of a subject for a particular class. The generated FC data is combined with the original training set, and the augmented data is used to train a downstream FC-based classifier to discriminate between brain disorder and healthy control groups. We consider three types of widely-used FC classifiers: support vector machine (SVM), convolutional neural network (CNN) and BrainNetCNN \cite{Jeremy2017}.


\section{An Application to fMRI FC Data Augmentation for Major Depression Identification}

In this section, we present experimental evaluation of the proposed GR-SPD-GAN model for generating realistic FC data, and its usefulness for data augmentation for MDD classification on a large MDD fMRI dataset.

\vspace{-0.05in}
\subsection{Data Acquisition and Pre-processing}

We used a resting-state fMRI dataset of 227 healthy controls (HC) and 250 Major Depressive Disorder (MDD) patients from the open-access REST-meta-MDD Consortium database \cite{Yan2019}. The MDD patients had a Hamilton Depression Rating Scale (HAMD) scores of $\geq 8$. The data were acquired using a Siemens (Tim Trio 3T) scanner (TR/TE = 2000/30 ms, 3mm slice thickness). The data were pre-processed using Data Processing Assistant for Resting-State fMRI (DPARSF) \cite{Yan2010}. See \cite{Yan2019} for more details. We used the Automated Anatomical Labeling (AAL) atlas to obtain brain parcellation into 116 region-of-interests (ROIs), including cortical and subcortical areas, and extracted mean time series of $232$ time points for each ROI. We finally estimated a $116 \times 116$ FC matrix for each subject, based on Pearson’s correlations between ROIs. We computed the Fréchet mean $\bm y$ of the brain FCs of each class (HC and MDD) by \eqref{eqn:frechet_mean}.

\vspace{-0.05in}
\subsection{Implementation details}

\subsubsection{Model Architecture and Training}
The $G$ and $D$ of the proposed GR-SPD-GAN are convolutional neural networks with multiple upsampling and downsampling blocks. The generator $G$ takes the concatenation of a 100-dimensional random vector sampled from a standard Gaussian distribution, and a 2-dimensional one-hot vector that encodes the class label of the brain networks (MDD or HC) as inputs. This vector is mapped to one fully-connected (dense) layer of 61,952 outputs, and four convolutional upsampling blocks with 256, 128, 64, and 1 output channels. Each upsampling block consists of the nearest-neighbor upsampling followed by a $2 \times 2$ stride convolution (Conv) with LeakyReLU activation function. We have a final dense layer with hyperbolic tangent activation to produce a vector, which is then reshaped to a symmetric $116 \times 116$ matrix as the final output of the GR-SPD-GAN generator. The discriminator takes a $116 \times 116$ FC matrix as input. It consists of three downsampling blocks with  256, 128, 64 output channels, where each block is a $2 \times 2$ stride Conv layer followed by batch normalization and and LeakyReLU activation. The last dense layer takes both the outputs of the downsampling layers and the class label, and uses the sigmoid activation function to output the probability of a FC sample being sampled from the true data distribution. We trained the GR-SPD-GAN with learning rate of 0.0001 using the Adam optimizer \cite{Kingma2014}, 100 training epochs, and a mini-batch size of 32. The hyper-parameters in \eqref{eqn:loss-GR-SPDGAN} are empirically determined as $\alpha_1 = 0.8, \alpha_2 = 0.8, \alpha_3 = 0.8, \alpha_4 = 1, \alpha_5 = 1$ from a range of parameters, which gave the optimal performance.

\subsubsection{Methods for Comparison}
We benchmark the performance of our proposed method with several standard and state-of-the-art GAN-based generative models, including the 1-dimensional deep convolutional GAN (1D-DCGAN), 2-dimensional DCGAN (2D-DCGAN) \cite{Radford2015}, WGAN \cite{Martin2017}, and WGAN-GP \cite{Ishaan2017}. The generators of all competing models are composed of three stacked upsampling blocks, including convolutional transpose and batch normalization layers. All of the GANs take brain FC matrices as inputs except the Vanilla-GAN \cite{Ian2014} and 1D-DCGAN which take the vectorized FC for training. The discriminator/critic in the GANs are composed of two stacked downsampling convolutional blocks where the activation function of the last fully-connected layer is defined as sigmoid/linear, and batch normalization layers are excluded in WGAN-GP. The discriminator of Vanilla-GAN uses fully-connected layers instead of convolutional blocks.

\begin{algorithm}[!t]
\caption{SPD data generation by GR-SPD-GAN}\label{alg:fca}
\begin{algorithmic}[1]
\REQUIRE  $G$, Generator trained with Algorithm \ref{alg:grmacwg};
        $\bm x$, training data;
        $c$, FC class condition.
        

\STATE Sample $\{ \bm z\} \sim \mathbb{P}_z$ random noise samples from noise prior

\STATE Generate $\tilde{\bm{v}} = G((\bm z,c))$, points on the tangent space of the manifold $\mathcal{M}$

\STATE Generate $\bm {\tilde x} = \exp_{\bm y}(\tilde{\bm{v}})$, by mapping the generated points $\tilde{\bm{v}}$ to the manifold $\mathcal{M}$ at reference point ${\bm y}$ using Eq\eqref{eqn:airm_exp}

\STATE Concatenate $\bm {\tilde x}$ and $\bm x$ as augmented training dataset

\end{algorithmic}
\end{algorithm}

\begin{figure*}[t!]
\centerline{\includegraphics[width=\textwidth]{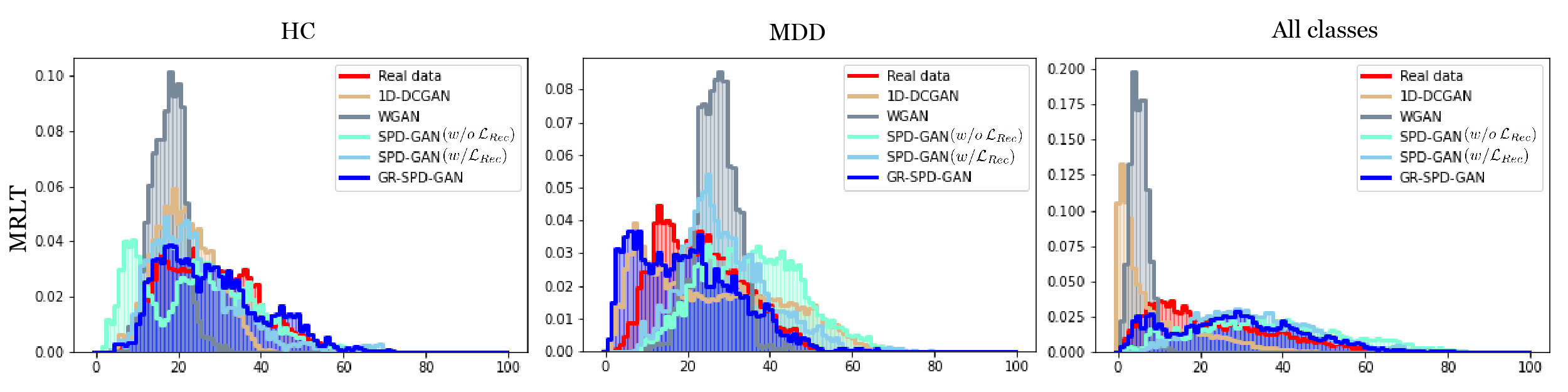}}
\vspace{-0.3cm}
\caption{Comparison of MRLT on the real and synthetic FC data distributions generated by 1D-DCGAN, WGAN, and SPD-GAN variants.}
\label{mrlt_results}
\vspace{-0.1in}
\end{figure*}

\begin{figure}[t!]
{\includegraphics[width=\columnwidth]{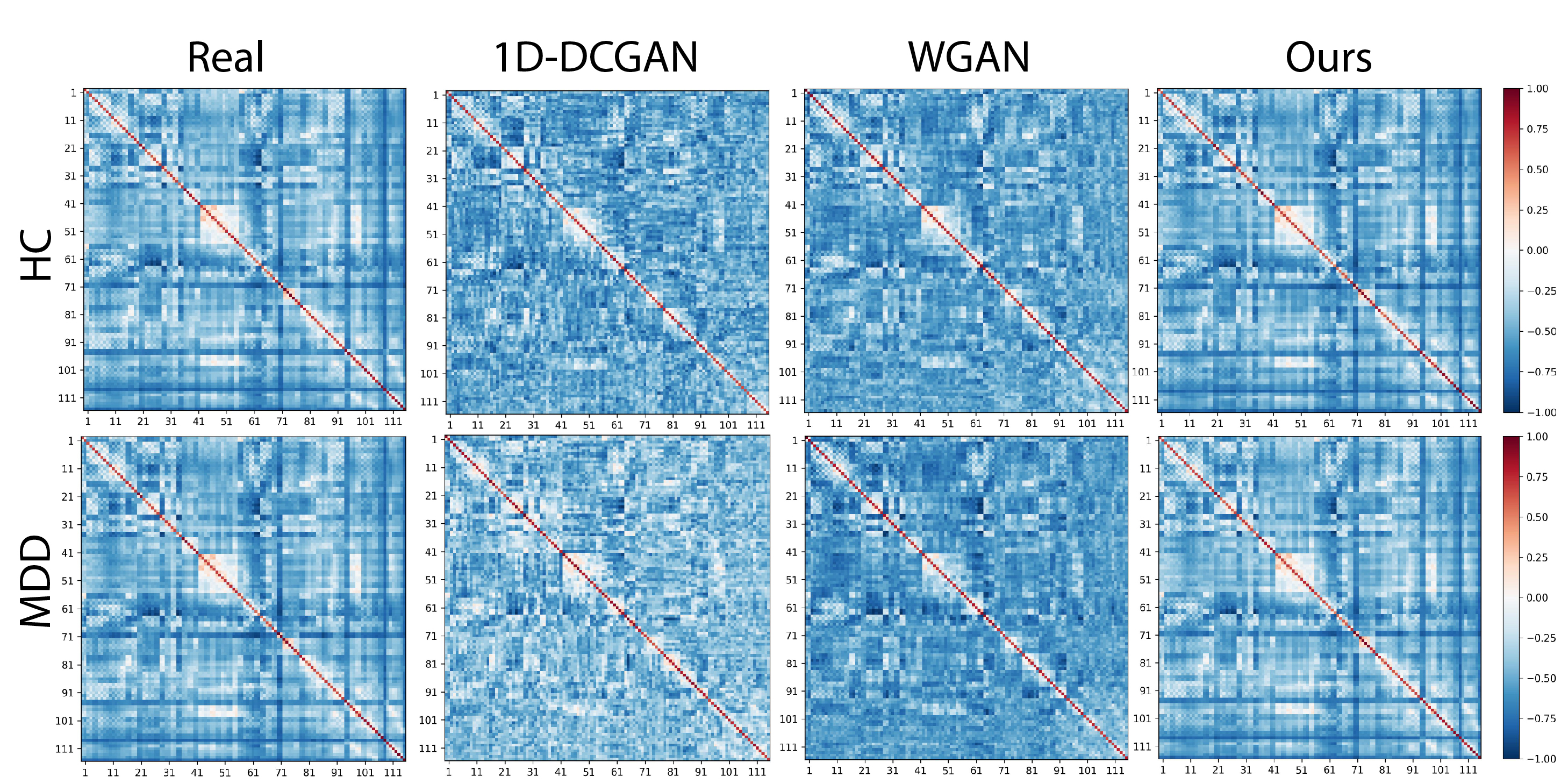}}
\caption{Ground-truth fMRI-derived FC matrices for MDD and HC (left). Generated samples by 1D-DCGAN, WGAN, and our GR-SPD-GAN.}
\label{fig:comparison}
\vspace{-0.15in}
\end{figure}

\subsubsection{Performance Measures}

We use the recently proposed geometry score (GS) \cite{Khrulkov2018} which provides both the qualitative and quantitative means to assess the quality of generated samples by GANs. GS compares the topological properties of the underlying real data manifold and the generated one, with a lower value indicating a better match. It is agnostic to the type of data, and is shown to be more expressive in capturing various failure modes of GANs compared to the conventional inception score and Fréchet inception distance. CS is computed based on mean relative living times (MRLT)
\begin{equation*}
\text{CS}(\bm{X}_1, \bm{X}_2)  = \sum^{i_{max}-1}_{i=0} (\text{MRLT}(i, 1,  \bm{X}_1) - \text{MRLT}(i, 1, \bm{X}_2))^2
\end{equation*}
$\text{MRLT}(i, k, \bm{X})$ can be interpreted as a probability distribution (over non-negative integers $i$) that defines the certainty about the correct number of $k$-dimensional holes (or connected components) in the underlying manifold of a dataset $\bm{X}$ on average. As in \cite{Khrulkov2018}, we used $k = 1$ to study the first homology of datasets.

%
%


\subsubsection{FC Augmentation \& Classification}
We assess the usefulness of the generated FC data from GR-SPD-GAN and the baseline generative methods via data augmentation for classifying MDD/healthy control. We trained the downstream FC classifiers on the augmented training data with different amounts of generated data, i.e. $\times1$, $\times2$ and $\times3$ multiple of the size of the original training set. We applied the nested-stratified 5-fold cross-validation (CV) \cite{Pereira2009} data partitioning scheme to evaluate the FC classification performance. Outer-folds consist of training and testing sets, where the training set was further split into 5-inner-folds training and validation sets. For each outer-fold, a GAN model was trained using the real training set to generate a fold-dependent synthetic dataset. Synthetic data were then combined with the real training data of inner-folds for the classifier training and hyper-parameters tuning. We used the Optuna library \cite{Akiba2019} to determine the best hyper-parameters of each classifier. The test set performance is evaluated based on the selected best models using accuracy, recall, precision, F1-score, and receiver operating characteristic (ROC).

The hyper-parameters search range, model configuration and the training of the three FC classifiers considered are: 
\begin{enumerate}[]
\item \textit{SVM}: SVM has been used to classify brain FC with reasonably good accuracy. We trained SVM on flattened (vectorized) brain FCs. 
The ranges considered in the hyper-parameter tuning: kernel function in $['rbf', 'linear']$, regularization parameter $C$ from $-5$ to $15$, and gamma value from $-15$ to $5$. 
\item \textit{CNN}: We adopt CNN with stacked convolutional blocks, each block consists of Conv2D, BatchNorm, and max-pooling (MaxPool) layers. Here, the correlation-based FC matrices were used directly as inputs to the CNN. The convolutional blocks learn high-level spatial FC which are then fed into the fully-connected and Softmax layers for classification.
Hyper-parameter search space: batch size from $5$ to $32$, number of convolutional layers from $1$ to $3$, kernel size in $ [10, 16, 24, 32]$, number of fully-connected layers from $1$ to $3$.
\item \textit{BrainNetCNN}: BrainNetCNN is a specially-designed CNN for brain connectivity with special convolutional kernels to preserve the brain network structure. It consists of three types of layers: edge-to-edge, edge-to-node and node-to-graph convolutional layers to capture topological relationships between brain network edges. Hyper-parameter search space: L2 regularization weight decay from $1e{-8}$ to $1e{-2}$, scheduler learning rate reduce factor from $0.1$ to $0.9$, batch size from $5$ to $16$.
\end{enumerate}
We implemented all DL models for FC data generation and classification based on the Tensorflow \cite{Abadi2016} framework on a PC with a single GPU-NVIDIA Quadro P5000 16 GB.


\begin{table}[!t]
\caption{Synthetic FC data evaluation with geometry score $\times 10^3$. } 
\label{tab:geometryscore}
\begin{tabularx}{\columnwidth}{llXXX}
\hline\hline
\textbf{}& Methods & HC & MDD & All\\
\hline
\multirow{5}{*}{Competing} 
    & Vanilla-GAN & 171.0 & 162.4 & 105.7  \\
    & 1D-DCGAN & 9.8 & 10.1 & 54.6 \\
    & 2D-DCGAN & 116.0 & 56.2 & 69.7 \\ 
    & WGAN & 38.4 & 29.9 & 47.5 \\  
    & WGAN-GP & 43.8 & 108.2 & 57.1 \\
 \cline{1-5}
\multirow{3}{*}{Ours} 
    & SPD-GAN (w/o $\mathcal{L}_{rec}$) & 8.7 & 15.1 & 8.4 \\
    & SPD-GAN (w/ $\mathcal{L}_{rec}$) & 3.9 & 8.5 & 10.5 \\   
    & GR-SPD-GAN & \textbf{1.7} & \textbf{5.2} & \textbf{4.2} \\
    \hline\hline
\end{tabularx}
\vspace{-0.1in}
\end{table}

\begin{table*}[!ht]

\caption{MDD classification performance of different FC classifiers trained on the original training set (real) and augmented datasets with an increasing amount of synthesized FC data using the proposed GR-SPD-GAN and other GAN-based generative models. Results are averages (standard deviations) of performance measures over 5-fold cross-validation.}
\label{tab:results}
\resizebox{\textwidth}{!}{
\begin{tabular}{cccccccc}
\hline\hline
Classifier	& GAN Type	& Train set	& Accuracy	& Recall	& Precision & F1-Score & ROC\\
\hline
\multirow{19}{*}{SVM} 
    & - & Real & $62.79 \pm 4.02 $ & $62.79 \pm 4.02 $ & $63.08 \pm 4.06 $ & $62.63 \pm 3.95 $ & $62.75 \pm 3.90 $ \\ 
    \cline{2-8}
& \multirow{3}{*}{Vanilla-GAN}  
    & Real + Synth. $1\times$ & $65.03 \pm 1.84$ & $65.03 \pm 1.84 $ & $65.36 \pm 2.30 $ & $64.76 \pm 1.50 $ & $64.82 \pm 1.63 $ \\
    & & Real + Synth. $2\times$ & $64.98 \pm 1.79$ & $64.98 \pm 1.79$ & $65.32 \pm 2.24 $ & $64.72 \pm 1.46 $ & $64.77 \pm 1.59 $ \\
    & & Real + Synth. $3\times$ & $65.15 \pm 1.74 $ & $65.15 \pm 1.74 $ & $65.48 \pm 2.19 $ & $64.90 \pm 1.41 $ & $64.95 \pm 1.53 $ \\
    \cline{2-8}
& \multirow{3}{*}{1D-DCGAN} 
    & Real + Synth. $1\times$ & $65.82 \pm 3.31 $ & $65.82 \pm 3.31 $ & $66.13 \pm 3.65 $ & $65.64 \pm 3.17 $ & $65.67 \pm 3.26 $ \\
    & & Real + Synth. $2\times$ &  $66.03 \pm 3.48 $ & $66.03 \pm 3.48 $ & $66.29 \pm 3.71 $ & $65.91 \pm 3.37 $ & $65.95 \pm 3.43 $ \\
    & & Real + Synth. $3\times$ & $65.78 \pm 3.02 $ & $65.78 \pm 3.02 $ & $66.00 \pm 3.18 $ & $65.68 \pm 2.94 $ & $65.71 \pm 2.98 $ \\
    \cline{2-8}
& \multirow{3}{*}{2D-DCGAN} 
    & Real + Synth. $1\times$ & $65.36 \pm 3.15 $ & $65.36 \pm 3.15 $ & $65.66 \pm 3.46 $ & $65.19 \pm 3.00 $ & $65.24 \pm 3.07 $ \\
    & & Real + Synth. $2\times$ & $65.07 \pm 2.90 $ & $65.07 \pm 2.90 $ & $65.35 \pm 3.08 $ & $64.94 \pm 2.76 $ & $65.00 \pm 2.79 $ \\
    & & Real + Synth. $3\times$ & $65.15 \pm 2.79 $ & $65.15 \pm 2.79 $ & $65.50 \pm 3.18 $ & $64.93 \pm 2.61 $ & $64.99 \pm 2.72 $ \\
    \cline{2-8}
& \multirow{3}{*}{WGAN} 
    & Real + Synth. $1\times$ & \boldmath{$66.28 \pm 3.59 $} & \boldmath{$66.28 \pm 3.59 $} & \boldmath{$66.52 \pm 3.72 $} & \boldmath{$66.19 \pm 3.50 $} & \boldmath{$66.23 \pm 3.53 $} \\
    & & Real + Synth. $2\times$ & $65.40 \pm 3.32 $ & $65.40 \pm 3.32 $ & $65.61 \pm 3.43 $ & $65.31 \pm 3.23 $ & $65.34 \pm 3.23 $ \\
    & & Real + Synth. $3\times$ & $66.03 \pm 3.28 $ & $66.03 \pm 3.28 $ & $66.28 \pm 3.45 $ & $65.92 \pm 3.19 $ & $65.96 \pm 3.25 $ \\
    \cline{2-8}
& \multirow{3}{*}{WGAN-GP} 
    & Real + Synth. $1\times$ & $63.42 \pm 3.00 $ & $63.42 \pm 3.00 $ & $63.58 \pm 2.99 $ & $63.31 \pm 2.89 $ & $63.31 \pm 2.83 $ \\
    & & Real + Synth. $2\times$ & $63.59 \pm 2.45 $ & $63.59 \pm 2.45 $ & $63.74 \pm 2.47 $ & $63.44 \pm 2.33 $ & $63.43 \pm 2.30 $ \\
    & & Real + Synth. $3\times$ & $63.51 \pm 2.61 $ & $63.51 \pm 2.61 $ & $63.64 \pm 2.63 $ & $63.37 \pm 2.50 $ & $63.35 \pm 2.46 $ \\
    \cline{2-8}
& \multirow{3}{*}{GR-SPD-GAN} 
    & Real + Synth. $1\times$ & $76.68 \pm 1.76$ & $76.68 \pm 1.76$ & $77.10 \pm 1.66 $ & $76.59 \pm 1.74 $ & $76.63 \pm 1.58 $ \\
    & & Real + Synth. $2\times$ & \boldmath{ $77.14 \pm 1.74$} & \boldmath{$77.14 \pm 1.74$} & \boldmath{$77.57 \pm 1.59 $} & \boldmath{$77.06 \pm 1.73 $} & \boldmath{ $77.14 \pm 1.58 $} \\
    & & Real + Synth. $3\times$ & \boldmath{ $77.14 \pm 1.74$} & \boldmath{$77.14 \pm 1.74$} & \boldmath{$77.57 \pm 1.59 $} & \boldmath{$77.06 \pm 1.73 $} & \boldmath{ $77.14 \pm 1.58 $} \\
\hline
\multirow{19}{*}{CNN} 
    & - & Real & $60.99 \pm 3.21 $ & $60.99 \pm 3.21 $ & $61.40 \pm 2.98 $ & $60.51 \pm 3.40 $ & $60.84 \pm 3.14 $ \\ 
    \cline{2-8}
& \multirow{3}{*}{Vanilla-GAN}  
    & Real + Synth. $1\times$ & $60.57 \pm 4.96$ & $60.57 \pm 4.96 $ & $61.78 \pm 7.01 $ & $59.61 \pm 4.49 $ & $60.06 \pm 4.81 $ \\
    & & Real + Synth. $2\times$ & $63.09 \pm 5.78 $ & $63.09 \pm 5.78 $ & $64.62 \pm 5.01 $ & $61.84 \pm 6.89 $ & $63.11 \pm 5.58 $ \\ 
    & & Real + Synth. $3\times$ & $65.00 \pm 3.73 $ & $65.00 \pm 3.73 $ & $65.35 \pm 3.93 $ & $64.69 \pm 3.71 $ & $64.76 \pm 3.75 $ \\
    \cline{2-8}
& \multirow{3}{*}{1D-DCGAN} 
    & Real + Synth. $1\times$ & $66.86 \pm 3.67$ & $66.86 \pm 3.67$ & $67.03 \pm 3.73$ & $66.69 \pm 3.72$ & $66.67 \pm 3.75$ \\
    & & Real + Synth. $2\times$ & $63.93 \pm 4.83 $ & $63.93 \pm 4.83 $ & $64.41 \pm 5.03 $ & $63.38 \pm 4.92 $ & $63.56 \pm 4.90 $ \\
    & & Real + Synth. $3\times$ & \boldmath{$68.32 \pm 5.12 $} & \boldmath{$68.32 \pm 5.12$} & \boldmath{$69.06 \pm 5.32 $} & \boldmath{$67.85 \pm 5.16 $} & \boldmath{$67.96 \pm 5.03 $} \\
    \cline{2-8}
    & \multirow{3}{*}{2D-DCGAN} 
    & Real + Synth. $1\times$  & $62.87 \pm 6.52$ & $62.87 \pm 6.52$ & $63.23 \pm 7.06$ & $62.17 \pm 6.65$ & $62.29 \pm 6.59$ \\
    & & Real + Synth. $2\times$ & $62.04 \pm 3.03$  & $62.04 \pm 3.03$ & $62.40 \pm 2.85$ & $61.53 \pm 3.41$ & $61.81 \pm 3.15$ \\
    & & Real + Synth. $3\times$  & $65.40 \pm 2.90$ & $65.40 \pm 2.90$ & $66.09 \pm 2.88$ & $64.87 \pm 3.02$ & $65.25 \pm 2.93$ \\
\cline{2-8}
& \multirow{3}{*}{WGAN} 
    & Real + Synth. $1\times$ & $ 64.15 \pm 2.62 $ & $ 64.15 \pm 2.62 $ & $ 64.35 \pm 2.61 $ & $ 64.07 \pm 2.65 $ & $ 64.14 \pm 2.64 $ \\
    & & Real + Synth. $2\times$ & $ 56.82 \pm 4.79 $ & $ 56.82 \pm 4.79 $ & $ 57.41 \pm 5.32 $ & $ 56.25 \pm 5.03 $ & $ 56.93 \pm 5.08 $ \\
    & & Real + Synth. $3\times$ & $ 63.52 \pm 4.19 $ & $ 63.52 \pm 4.19 $ & $ 64.01 \pm 4.25 $ & $ 62.88 \pm 4.80 $ & $ 63.29 \pm 4.60 $ \\
    \cline{2-8}
& \multirow{3}{*}{WGAN-GP} 
    & Real + Synth. $1\times$ & $ 61.02 \pm 4.01 $ & $ 61.02 \pm 4.01 $ & $ 61.76 \pm 3.96 $ & $ 60.71 \pm 4.20 $ & $ 61.30 \pm 3.88  $ \\
    & & Real + Synth. $2\times$ & $ 62.87 \pm 6.99 $ & $ 62.87 \pm 6.99 $ & $ 63.54 \pm 6.75 $ & $ 62.42 \pm 7.25 $ & $ 62.99 \pm 6.70 $ \\
    & & Real + Synth. $3\times$ & $ 63.49 \pm 6.36$ & $ 63.49 \pm 6.36 $ & $ 64.11 \pm 6.33 $ & $ 63.00 \pm 6.52 $ & $ 63.38 \pm 6.26 $ \\
    \cline{2-8}
& \multirow{3}{*}{GR-SPD-GAN} 
    & Real + Synth. $1\times$ & $70.64 \pm 3.17$ & $70.64 \pm 3.17 $ & $70.72 \pm 3.20 $ & $70.61 \pm 3.15 $ & $70.56 \pm 3.16 $ \\
    & & Real + Synth. $2\times$ & $70.01 \pm 2.29 $ & $70.01 \pm 2.29 $ & $70.18 \pm 2.36 $ & $70.01 \pm 2.27 $ & $70.06 \pm 2.33 $ \\ 
    & & Real + Synth. $3\times$ & \boldmath{$70.86 \pm 1.47 $} & \boldmath{$70.86 \pm 1.47 $} & \boldmath{$71.36 \pm 1.67 $} & \boldmath{$70.78 \pm 1.43 $} & \boldmath{$70.99 \pm 1.50 $} \\
\hline
\multirow{19}{*}{BrainNetCNN} 
    & - & Real & $58.90 \pm 2.98$  & $58.90 \pm 2.98$ & $59.56 \pm 2.74$ & $58.39 \pm 3.09$ & $59.00 \pm 2.56$ \\
    \cline{2-8}
& \multirow{3}{*}{Vanilla-GAN}  
    & Real + Synth. $1\times$ & $56.90 \pm 1.66$  & $56.90 \pm 1.66$ & $56.40 \pm 2.86$ & $53.68 \pm 3.31$ & $56.29 \pm 1.92$  \\
    & & Real + Synth. $2\times$ & $50.71 \pm 3.69$  & $50.71 \pm 3.69$ & $48.60 \pm 7.23$ & $46.74 \pm 5.18$ & $50.81 \pm 4.13$  \\
    & & Real + Synth. $3\times$ & $58.86 \pm 2.24$  & $58.86 \pm 2.24$ & $59.91 \pm 2.57$ & $57.64 \pm 1.96$ & $58.57 \pm 2.05$  \\
    \cline{2-8}
& \multirow{3}{*}{1D-DCGAN} 
    & Real + Synth. $1\times$ & $62.94 \pm 2.01$  & $62.94 \pm 2.01$ & $63.43 \pm 2.20$ & $62.23 \pm 2.68$ & $62.71 \pm 2.26$  \\
    & & Real + Synth. $2\times$ & $65.04 \pm 2.02$ & $65.04 \pm 2.02$ & $66.35 \pm 2.13$ & $64.12 \pm 2.10$ & $64.74 \pm 2.08$ \\
    & & Real + Synth. $3\times$ & $58.21 \pm 2.98$  & $58.21 \pm 2.98$ & $55.70 \pm 6.58$ & $52.86 \pm 4.14$ & $57.38 \pm 3.11$  \\
    \cline{2-8}
    & \multirow{3}{*}{2D-DCGAN} 
    & Real + Synth. $1\times$  & $60.78 \pm 4.98$ & $60.78 \pm 4.98$ & $61.30 \pm 5.34$ & $60.01 \pm 5.00$ & $60.33 \pm 5.03$ \\
    & & Real + Synth. $2\times$  & $61.41 \pm 2.59$ & $61.41 \pm 2.59$ & $61.99 \pm 3.73$ & $62.18 \pm 3.29$ & $61.04 \pm 2.79$ \\
    & & Real + Synth. $3\times$  & $62.88 \pm 4.99$  & $62.88 \pm 4.99$ & $63.12 \pm 5.02$ & $62.48 \pm 5.25$ & $62.67 \pm 5.15$ \\
    \cline{2-8}
& \multirow{3}{*}{WGAN} 
    & Real + Synth. $1\times$ & $ 64.98 \pm 5.54 $ & $ 64.98 \pm 5.54  $ & $ 65.19 \pm 5.34 $ & $ 64.86 \pm 5.61 $ & $ 64.95 \pm 5.39 $ \\
    & & Real + Synth. $2\times$ & $ 60.59 \pm 1.81 $ & $ 60.59 \pm 1.81 $ & $ 60.89 \pm 1.96 $ & $ 60.35 \pm 1.78 $ & $ 60.53 \pm 1.84  $ \\
    & & Real + Synth. $3\times$ & $ 61.83 \pm 3.03 $ & $ 61.83 \pm 3.03 $ & $ 62.27 \pm 3.29$ & $ 61.44 \pm 2.70 $ & $ 61.58 \pm 2.73 $ \\
    \cline{2-8}
& \multirow{3}{*}{WGAN-GP} 
    & Real + Synth. $1\times$ & \boldmath{$ 66.02 \pm 4.25 $} & \boldmath{$ 66.02 \pm 4.25 $} & \boldmath{$ 66.22 \pm 4.24 $} & \boldmath{$ 65.93 \pm 4.20 $} & \boldmath{$ 65.95 \pm 4.13 $} \\
    & & Real + Synth. $2\times$ & $ 64.76 \pm 4.25 $ & $ 64.76 \pm 4.25  $ & $ 65.67 \pm 4.08 $ & $ 64.23 \pm 4.52 $ & $ 64.73 \pm 4.14  $ \\
    & & Real + Synth. $3\times$ & $ 64.56 \pm 3.18 $ & $ 64.56 \pm 3.18  $ & $ 64.78 \pm 3.17 $ & $ 64.38 \pm 3.15 $ & $ 64.41 \pm 3.08  $ \\
  \cline{2-8}
& \multirow{3}{*}{GR-SPD-GAN} 
    & Real + Synth. $1\times$ & $ 71.26 \pm 3.90 $  & $ 71.26 \pm 3.90 $ & $ 72.17 \pm 4.23 $ & $ 70.95 \pm 3.84 $ & $ 71.08 \pm 3.67 $  \\
    & & Real + Synth. $2\times$ & $ 71.47 \pm 3.72 $  & $ 71.47 \pm 3.72 $ & $ 71.71 \pm 3.91 $ & $ 71.36 \pm 3.64 $ & $ 71.29 \pm 3.61 $  \\
    & & Real + Synth. $3\times$ & \boldmath{$74.42 \pm 1.70$}  & \boldmath{$74.42 \pm 1.70$} & \boldmath{$74.27 \pm 1.59$} & \boldmath{$74.35 \pm 1.69$} & \boldmath{$74.39 \pm 1.56$}  \\
\hline\hline

\end{tabular}
}
\vspace{-0.3cm}
\end{table*}

\subsection{Results}
\subsubsection{Quality of Synthetic FC Data}

\begin{table*}[!htbp]
\caption{Effect of incorporating different component losses in the training of GR-SPD-GAN to augment data for FC classification.} 
\label{tab:ablation_study}
\resizebox{\textwidth}{!}{
\begin{tabular}{c|ccc|c|ccccc}
\hline\hline
Ours	& $\mathcal{L}_{adv}$	& $\mathcal{L}_{Rec}$	& $\mathcal{L}_{Graph}$	& Classifiers	& Accuracy & Recall & Precision & F1-Score & ROC\\
\hline
\multirow{3}{*}{SPD-GAN (w/o $\mathcal{L}_{Rec}$)} 
    & \multirow{3}{*}{\cmark} & \multirow{3}{*}{\xmark} & \multirow{3}{*}{\xmark}  & SVM & $76.09 \pm 3.31$  & $76.09 \pm 3.31$ & $76.34 \pm 3.35$ & $76.02 \pm 3.30$ & $76.00 \pm 3.26$ \\
    &  &  &   & CNN & $63.95 \pm 4.09$  & $63.95 \pm 4.09$ & $64.23 \pm 4.41$ & $63.53 \pm 4.36$ & $63.67 \pm 4.44$ \\
    &  &  &   & BrainNetCNN & $ 72.32 \pm 3.56 $  & $ 72.32 \pm 3.56 $ & $ 72.98 \pm 2.92 $ & $ 72.13 \pm 3.75 $ & $ 72.39 \pm 3.35 $ \\
    \hline
\multirow{3}{*}{SPD-GAN (w/ $\mathcal{L}_{Rec}$)} 
    & \multirow{3}{*}{\cmark} & \multirow{3}{*}{\cmark} & \multirow{3}{*}{\xmark}  & SVM & $75.83 \pm 3.86$ & $75.83 \pm 3.86$ & $76.00 \pm 3.96 $ & $75.76 \pm 3.85 $ & $75.68 \pm 3.83 $ \\
    &  &  &   & CNN & $ 68.54 \pm 4.60 $  & $ 68.54 \pm 4.60 $  & $ 68.72 \pm 2.60 $ & $ 68.33 \pm 2.72 $ & $ 68.28 \pm 2.75 $ \\
    &  &  &   & BrainNetCNN & $ 72.74 \pm 2.95 $  & $ 72.74 \pm 2.95 $ & $ 73.99 \pm 3.58 $ & $ 72.28 \pm 3.15 $ & $ 72.38 \pm 2.97 $ \\
    \hline
\multirow{3}{*}{GR-SPD-GAN} 
    & \multirow{3}{*}{\cmark} & \multirow{3}{*}{\cmark} & \multirow{3}{*}{\cmark}  & SVM &  \boldmath{ $77.14 \pm 1.74$} & \boldmath{$77.14 \pm 1.74$} & \boldmath{$77.57 \pm 1.59 $} & \boldmath{$77.06 \pm 1.73 $} & \boldmath{ $77.14 \pm 1.58 $} \\
    &  &  &   & CNN & \boldmath{$70.86 \pm 1.47 $} & \boldmath{$70.86 \pm 1.47 $} & \boldmath{$71.36 \pm 1.67 $} & \boldmath{$70.78 \pm 1.43 $} & \boldmath{$70.99 \pm 1.50 $} \\
    &  &  &   & BrainNetCNN & \boldmath{$74.42 \pm 1.70$}  & \boldmath{$74.42 \pm 1.70$} & \boldmath{$74.27 \pm 1.59$} & \boldmath{$74.35 \pm 1.69$} & \boldmath{$74.39 \pm 1.56$} \\
\hline\hline
\end{tabular}
}
\end{table*}

Table.~\ref{tab:geometryscore} shows the geometry scores of the FC samples generated by different GANs. The proposed SPD-GANs obtained significantly better performance than classical GANs, with GR-SPD-GAN achieving the lowest value of geometry score for HC, MDD, and all classes, suggesting the highest similarity between the true and synthetic data distributions and its ability to avoid mode collapse. This is further confirmed by the MRLTs in Fig.~\ref{mrlt_results} that provides a visual quality of generated results. The synthetic data distributions by GR-SPD-GAN have the closest resemblance to the real data distributions for HC, MDD, and all classes, in contrast to the obvious deviations from the ground-truth for the 1D-DCGAN and WGAN. 
This suggest that generation of synthetic FC is better performed on SPD manifold. The effectiveness of GR-SPD-GAN in synthesizing more realistic FC is owing to its generation mechanism that leverages on the right underlying geometry (the Riemannian manifold) of the correlation-based FC matrices as SPD objects, and thus preserving the interrelatedness of edges in the generated FC matrices. On the other hand, GANs traditionally designed for Euclidean-valued data fail to capture the correlated nature of edges in real FC matrices.
Among the SPD-GANs, inclusion of the reconstruction losses $\mathcal{L}_{rec}$ and graph regularizers $\mathcal{L}_{graph}$ in GR-SPD-GAN produce a better match to the real data distributions.
In Fig.~\ref{fig:comparison}, we compare qualitatively synthetic FC matrices generated by different GANs. Results shown are group averages for the MDD and HC groups. Again, we observe that our model generated more realistic samples that can preserve the overall connectivity patterns and fine details of edges in the real FC networks.


\subsubsection{Results for FC Classification}

Table.~\ref{tab:results} shows the FC classification results for data augmentation using different GANs. Data augmentation using GANs is generally beneficial, giving significant gains in classification performance over the baseline models without data augmentation. Among the GAN generators, using generated data from our GR-SPD-GAN is clearly the most conducive to improving the classification performance across all classifiers, followed by WGAN and DCGANs. This is evident from the substantially larger margin of increase in all performance measures compared to the other GANs. The better classification performance with data augmentation using GR-SPD-GAN is due to the high quality of the synthetic FC data with SPD structure intact as shown in results in Fig.~\ref{mrlt_results}, Fig.~\ref{fig:comparison} and Table.~\ref{tab:geometryscore}. In contrast, the standard GAN architectures which neglect the SPD geometry are limited by the problem of mode collapse, producing only a small variety of FC samples with many duplicates (modes). When used for data augmentation, this will give little improvements or even degrade the classification performance due to over-fitting of classification models on duplicated samples. We also note that results saturate when augmenting with synthetic data by more than 3 times of the original training set. Moreover, the variances of classification performance measures obtained by data augmentation using GR-SPD-GAN are substantially lower in all classifiers compared to other GANs. This indicates that FC classifiers trained on data generated by GR-SPD-GAN achieved consistently good performance across different experimental folds, again implying better generalizability of the classifiers due to augmentation with the better quality and higher diversity of the synthesized data.

\subsubsection{Ablation Study}

We conducted an ablation study to evaluate the effect of each loss component used in training of our GR-SPD-GAN on FC classification performance. We consider three variants of GR-SPD-GAN to augment FC data for MDD classification. Each model was trained by discarding one loss component of the full GR-SPD-GAN model: (i) SPD-GAN (w/o $\mathcal{L}_{rec}$) with only the adversarial loss $\mathcal{L}_{adv}$; (ii) SPD-GAN (w/ $\mathcal{L}_{rec}$) with added reconstruction losses $\mathcal{L}_{rec}$ on manifold and tangent space, (iii) GR-SPD-GAN with $\mathcal{L}_{adv}$, $\mathcal{L}_{rec}$ as well as the population graph regularizers $\mathcal{L}_{graph}$ on manifold and tangent space. Table \ref{tab:ablation_study} shows the classification results of using $3 \times$ augmented FC data generated from these SPD-GANs. The SPD-GAN models trained with and without the reconstruction losses yield comparable results. However, adding the population graph regularizers to guide the GR-SPD-GAN training leads to noticeable improvement in classification performance. This suggests the usefulness of this loss incorporating the inter-subject relationships to generate more realistic FC data, that can further enhance FC classification when used for data augmentation.

\section{Conclusion}


We introduced a new manifold-aware deep generative model, GR-SPD-GAN for SPD-matrix valued data generation, by exploiting the unique manifold geometry of SPD matrices in the adversarial training. It uses a generalization of distribution distance in Wasserstein GAN for manifold-valued data. We demonstrated its usefulness for synthesizing correlation matrices of neuroimaging data, a common representation of functional brain networks that can preserve its SPD structure, and thus taking into account the inter-related nature of all FC edges as a whole. By devising a conditional mechanism in this GAN that uses class labels to supervise generation, it allows us to generate FC samples residing on the SPD manifold, and according to different classes of brain networks (with brain disorder or healthy control). We incorporated additional population graph-based regularization terms that can improve the GAN training, by forcing the generator to respect the inter-subject similarity of the FC structure in the real data in order to generate high quality data. Qualitative and quantitative results on a MDD fMRI dataset show the superiority of our method in generating more realistic FC samples, significantly outperforming state-of-the-art GAN-based generators in terms of geometric score. When applied to augment FC data for connectome-based MDD identification, our method also provided the largest improvements in classification performance. To conclude, the proposed GR-SPD-GAN approach is the first to show the advantages of FC data generation on the SPD space rather than Euclidean geometry, which can find a wide range of other applications in FC analyses besides data augmentation for brain disorder classification. Nevertheless, our method focuses on generation of static brain networks. Future work could consider extensions to generate dynamic FC \cite{Hutchison2013,Ting2021,Ting2022}, potentially by mapping the dynamic FC sequences as trajectories on space of SPD matrices as in \cite{Dai2020}.

\bibliographystyle{IEEEbib}
\bibliography{References}

\end{document}